\documentclass[lettersize,journal]{IEEEtran}
\usepackage{amsmath,amsfonts}
\usepackage{algorithmic}
\usepackage{algorithm}
\usepackage{array}
\usepackage[caption=false,font=normalsize,labelfont=sf,textfont=sf]{subfig}
\usepackage{textcomp}
\usepackage{stfloats}
\usepackage{url}
\usepackage{verbatim}
\usepackage{graphicx}
\usepackage{cite}
\usepackage{hyperref}

\begin{document}

\title{\fontsize{24}{28}\selectfont Sparse Phased Array Optimization Using Deep Learning}


\author{
	\IEEEauthorblockN{David Lin Yi Lu\IEEEauthorrefmark{1}, Lior Maman\IEEEauthorrefmark{2}, Jackson Earls\IEEEauthorrefmark{1}, Amir Boag\IEEEauthorrefmark{2}, Pierre Baldi\IEEEauthorrefmark{1}}
	
\IEEEauthorblockA{\IEEEauthorrefmark{1} Department of Computer Science, 
	University of California, Irvine, Irvine, USA \\}

\IEEEauthorblockA{\IEEEauthorrefmark{2} School of Electrical and Computer Engineering, 
	Tel Aviv University, Tel Aviv, Isreal \\
	(lud19@uci.edu, maman@mail.tau.ac.il, jearls@uci.edu, boag@tauex.tau.ac.il, pfbaldi@ics.uci.edu)}
	}



\pagenumbering{gobble}

\maketitle

\begin{abstract}
Antenna arrays are widely used in wireless communication, radar systems, radio astronomy, and military defense to enhance signal strength, directivity, and interference suppression. We introduce a deep learning-based optimization approach that enhances the design of sparse phased arrays by reducing grating lobes. This approach begins by generating sparse array configurations to address the non-convex challenges and extensive degrees of freedom inherent in array design. We use neural networks to approximate the non-convex cost function that estimates the energy ratio between the main and side lobes. This differentiable approximation facilitates cost function minimization through gradient descent, optimizing the antenna elements' coordinates and leading to an improved layout. Additionally, we incorporate a tailored penalty mechanism that includes various physical and design constraints into the optimization process, enhancing its robustness and practical applicability. We demonstrate the effectiveness of our method by applying it to the ten array configurations with the lowest initial costs, achieving further cost reductions ranging from 411\% to 643\%, with an impressive average improvement of 552\%. By significantly reducing side lobe levels in antenna arrays, this breakthrough paves the way for ultra-precise beamforming, enhanced interference mitigation, and next-generation wireless and radar systems with unprecedented efficiency and clarity. The code for the method is available at \href{https://github.com/david5010/Optimization-of-Antenna-Arrays}{https://github.com/david5010/Optimization-of-Antenna-Arrays}.
\end{abstract}


\section{Introduction}
Phased array systems utilizing sparse configurations frequently encounter grating lobes, primarily due to limitations on the element count and beamwidth specifications that determine the antenna aperture size. These lobes often result from the periodic arrangements typical in conventional antenna arrays \cite{Amitay1972}. To address this, sparse arrays adopt aperiodic layouts. Nonetheless, designing these aperiodic arrays introduces significant challenges, such as an increased number of degrees of freedom to optimize and the substantial computational effort required to evaluate antenna radiation patterns across all main beam scan angles. Traditional techniques struggle to consistently identify global optima due to the complexity of the problem's landscape. The grid search method, though comprehensive, becomes impractical due to the myriad of possible configurations and the extensive computational resources needed.

Various strategies have been proposed to optimize thinned and sub-array configurations, employing advanced techniques to refine antenna design \cite{rocca2016unconventional, rocca2014ga, rocca2009sidelobe, rocca2009improved, haupt2010antenna, haupt1994thinned, haupt2005interleaved}. Additionally, prior methods focusing on sparse arrays have leveraged stochastic optimization techniques like simulated annealing to design aperiodic linear sparse arrays, aiming to minimize side lobe levels and optimize array configurations \cite{trucco1999stochastic}. While these methods are adaptable and capable of handling complex, non-linear optimization challenges, they tend to be computationally demanding and slow, particularly for large arrays.

In recent years, machine learning techniques, particularly neural networks, have been increasingly applied to optimize complex issues in electromagnetics (EM). These networks effectively model computationally intensive EM simulations \cite{4120272, 1262734} and intricate EM relationships \cite{4120274, liu2019microwaveintegratedcircuitsdesign} with high accuracy. Beamforming, a signal processing technique, has been a focus of numerous deep learning applications \cite{8880526, Lin_2020, 9495844, 10137449}. Rather than inputting a desired array pattern into the neural network \cite{10137449}, our model takes a given sparse array as input and iteratively refines its geometry through gradient descent on the array coordinates.

To tackle computational speed and complexity, we initially reduce the number of variables in the phased array scanning problem from five (the observation vector, the scanning vector, and the wave number) to two \cite{maman2023phase, maman2021beam, maman2024beam, maman2025}. Subsequently, we generate sparse array data using a sub-array method \cite{boag2018grating}, which lessens the degrees of freedom in the optimization process. From this data, we select promising array configurations and further refine them using a deep learning approach. By employing neural networks, we accurately approximate the complex, non-convex cost function of an array, enabling efficient minimization of the cost through gradient descent on the antenna array coordinates.

\section{Methods}
\subsection{Generation of Antenna Array Configurations}

In this section, we tackle the challenge of designing Active Electronically Scanned Arrays (AESA) by simplifying the AESA array factor (AF) characterization from five variables to two and outlining the data generation process for optimization. The array factor (AF) for a planar phased array antenna, with uniformly excited elements situated in the $yz$-plane, is defined as follows:

\begin{equation}
  U(k, AZ, EL, AZ_0, EL_0) = \sum_{n=1}^{N_e} e^{jk\mathbf{r}_n\cdot(\mathbf{\hat{r}-\mathbf{\hat{r}_0}}}) 
\end{equation}

Here, $k$ is the wave-number, $\mathbf{\hat{r}} = \cos(EL) \sin(AZ) \mathbf{\hat{y}}  + \sin(EL) \mathbf{\hat{z}} + \cos(EL) \cos(AZ) \mathbf{\hat{x}} $ represents the unit vector in the observation direction, $
\mathbf{\hat{r}}_0 = \cos(EL_0) \sin(AZ_0) \mathbf{\hat{y}} + \sin(EL_0) \mathbf{\hat{z}} + \cos(EL_0) \cos(AZ_0) \mathbf{\hat{x}}
$ indicates the main beam direction. Additionally, $EL$ and $AZ$ refer to the elevation and azimuth of the direction of observation, respectively, while $EL_0$ and $AZ_0$ relate to the main beam direction. It is noted that $EL = \pi/2 - \theta$ and $ AZ = \varphi$ are in the conventional spherical coordinates system. The $\mathbf{r}_n = y_n \mathbf{\hat{y}} + z_n \mathbf{\hat{z}}$ specifies the location of the $n$th element and $N_e$  represents the number of elements. By introducing a difference vector $\mathbf{u} =k(\mathbf{\hat{r}}-\mathbf{\hat{r}}_0)$, we redefine the array factor using only two variables:
\begin{equation}
    U(u_y, u_z) = \sum_{n=1}^{N_e} e^{j(y_n u_y+z_n u_z)}
\end{equation}$u_y$  and $u_z$ are the $y$ and $z$ components of $\mathbf{u}$. The domain of $\mathbf{u}$ can be determined numerically or analytically based on the main beam scan sector. Identifying the AF's support within the $u_y$, $u_z$ plane simplifies the evaluation over the entire domain of interest. This method reduces the resources required for storage and processing, facilitating easier characterization of the entire observation area across the complete scan sector in a single illustration.
 
To streamline the optimization process and manage the search domain effectively, we adopted the concept of local periodicity to generate data \cite{boag2018grating}. This principle aids in organizing and integrating radiating elements and their associated transmit/receive modules through a sub-array strategy. In this approach, the entire array is constructed from multiple sub-arrays, each designed to exhibit periodicity. This strategy effectively narrows down the design variables. Practically, the array is configured with periodic sub-arrays, each capable of having distinct periodicity along both axes. These sub-arrays are also designed with the flexibility to be rotated and repositioned within their designated areas. The array’s surface is segmented into various subdomains, each hosting a periodic sub-array. The primary design parameters for optimization include the periodicity along each axis, as well as the rotation and positional adjustments of each sub-array. This approach simplifies the complex task of array design by reducing the multitude of potential configurations to a structured set of variables. Fig. 1 illustrates an array configuration utilizing a periodic sub-array structure.

\begin{figure}[htbp]
\centerline{\includegraphics[width=\linewidth]{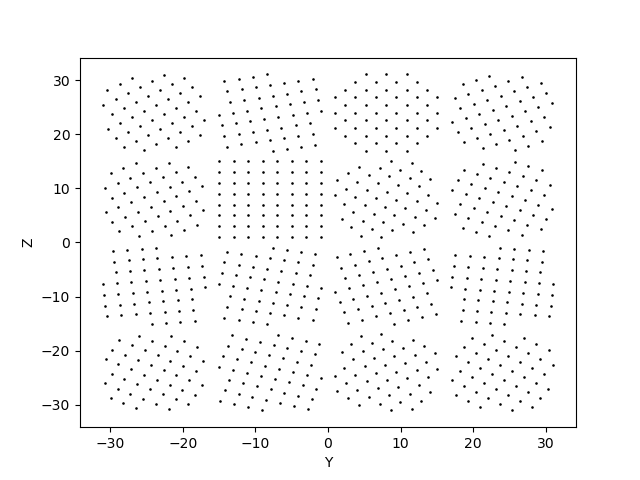}}
\caption{Example of a periodic sub-array configuration.}
\label{fig:SampleAnt1}
\end{figure}

\subsection{Cost Function}

The cost function evaluates the ratio of the main lobe (ML) to the side-lobe level (SLL) while adhering to the constraints
\begin{equation}
    f_{cost} = \text{--- }\frac{\sum_{i \in ML}|AF(u_y(i),u_z(i))|^{2p}}{\sum_{i \notin ML}|AF(u_y(i),u_z(i))|^{2p}}
\end{equation}where $p$ determines the norm type and $i$ is an index of samples in the $(u_y, u_z)$-plane.  For our data generation, we set $p=4$ to impose higher penalties on significant peaks, ensuring that excessively high sidelobes, such as grating lobes, do not appear outside the main lobe area.

\section{Optimization of Antenna Arrays using Neural Network Approximations}
Optimizing antenna array configurations is a formidable task, especially when aiming to minimize a complex and potentially non-differentiable cost function. The intricacy of the actual cost function demands considerable engineering efforts for direct optimization. To bypass these hurdles, we utilize neural networks as a strategic tool to accurately approximate the true cost function, acting as a surrogate. Neural networks are recognized for their universal approximation capability \cite{baldi2021deep, scarselli1998universal, patel2023neuralnetworktrainingnondifferentiable, DeVore_Hanin_Petrova_2021}. By nature, they provide a smooth and continuous function, which facilitates the use of gradient descent methods for optimization. The surrogate function is highly adaptable, as long as it provides an accurate estimate of the original cost function. This surrogate can adopt various forms, including feedforward neural networks (FNNs), convolutional neural networks (CNNs), or transformer variants, with the primary optimization process initiating after obtaining a reliable estimate. In our experiments, both an FNN and a Set Transformer \cite{pmlr-v97-lee19d} were evaluated as surrogate functions.

\subsection{Model Selection}
For the selection of optimal model architecture and hyperparameters, we utilize SHERPA, a hyperparameter optimization tool \cite{hertel2020sherpa}. We conduct a grid search for each adjustable parameter in both the Feedforward Neural Network (FNN) and Set Transformer models.

\subsubsection{Feedforward Neural Network (FNN) Model}
The FNN model architecture consists of four fully connected layers. The initial input layer is configured with a dimension of 2048, representing the Y and Z coordinates of 1024 array elements. This is followed by two hidden layers with dimensions of 20 and 12, respectively. ReLU activation functions are applied after every fully connected layer to introduce non-linearity into the network. The final output layer is equipped with a single neuron, tasked with predicting the cost associated with the specified antenna array geometry.

During the training phase, the FNN was subjected to 1000 epochs with a learning rate of \(1 \times 10^{-5}\) and a batch size of 128, employing the Adam optimizer to minimize the loss function.

We explore various configurations, including first and second hidden layer dimensions ranging from [4, 120] and [2, 40] neurons, respectively. Learning rates varied from \(1 \times 10^{-2}\) to \(1 \times 10^{-6}\), and batch sizes ranged from 16 to 256 to determine the most effective training parameters.

\begin{figure}[htbp]
\centerline{\includegraphics[width=\linewidth]{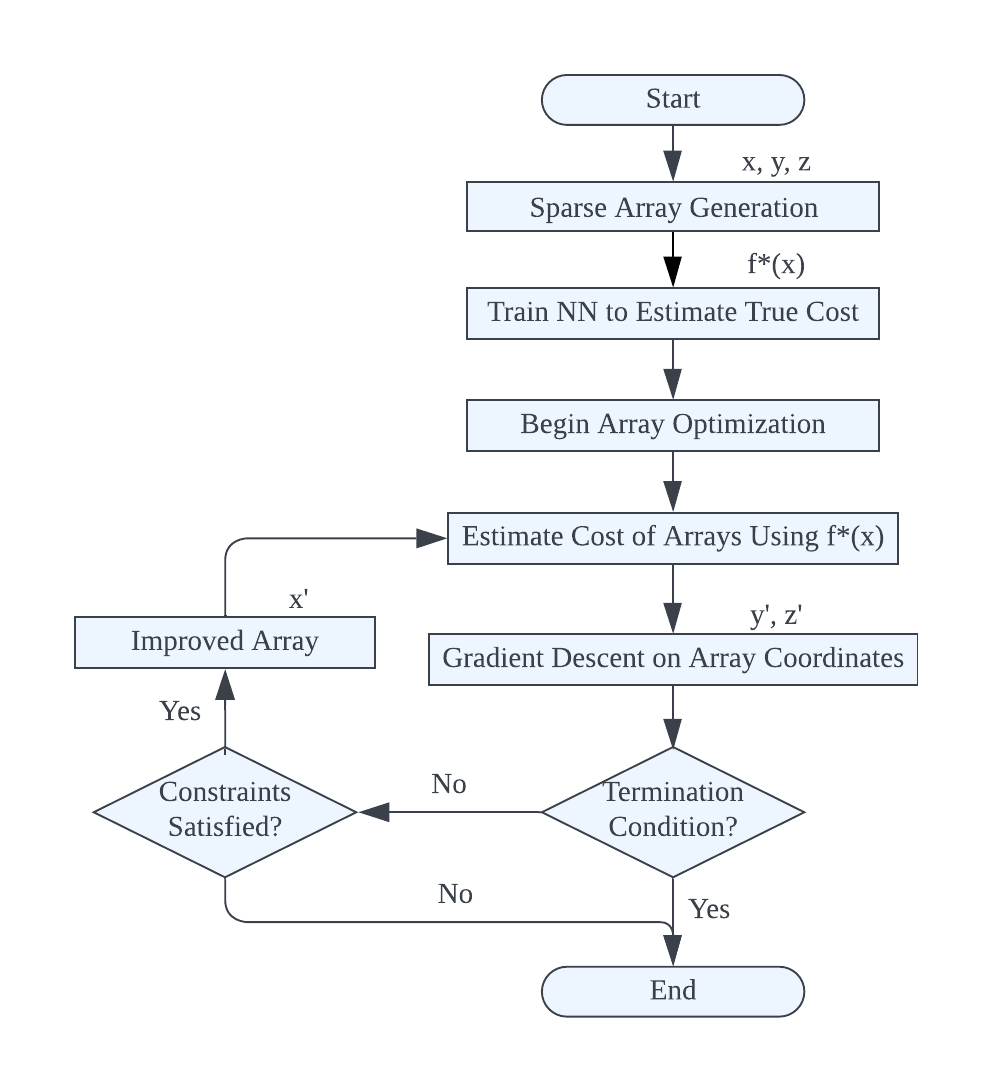}}
\caption{Flowchart illustrating iterative optimization process}
\label{fig:flowchart}
\end{figure}
\subsubsection{Set Transformer}
The Set Transformer is designed to handle set-structured data, as outlined by Lee et al. \cite{pmlr-v97-lee19d}. It features an encoder and a decoder configuration. The encoder is equipped with two permutation equivariant set attention blocks to maintain order invariance of the inputs, while the decoder incorporates a multi-head attention block for pooling, two additional set-attention blocks, and a fully connected output layer.

Both the encoder and decoder are structured with attention blocks that each have two heads, and the hidden layers within sustain a dimensionality of 32. The model processes inputs with a dimension of 2, corresponding to the Y and Z coordinates of each antenna element, and the output layer, akin to the FNN model, includes a single neuron that calculates the cost for the antenna array configuration.

The training of the Set Transformer was conducted over 1000 epochs with a learning rate of \(1 \times 10^{-3}\), a batch size of 64, using the Adam optimizer.

We evaluated different Set Transformer configurations by varying the hidden layer dimensions from 8 to 64 and the number of attention heads from 2 to 8. We explore learning rates between \(1 \times 10^{-2}\) and \(1 \times 10^{-6}\) and adjust batch sizes from 16 to 256 to determine optimal hyperparameters.

\subsection{Cost}
To evaluate the cost of an antenna array configuration, the neural network produces a real number, which spans from -40,000 to -0.7, with a lower cost signifying improved performance. Due to the broad spectrum of output values, we normalized these costs for training purposes and then utilized equation (\ref{eq:eq_saling}) to scale the training data. For the sake of interpretability, we apply reverse scaling to the outputs to restore the original cost values.
\begin{equation}
\label{eq:eq_saling}
    Y_\text{scaled} = \frac{Y - \mu_{y}}{\sigma_{y}}
\end{equation}

\begin{equation}
    Y = (Y_\text{scaled} \cdot \sigma_{y}) + \mu_{y}
\end{equation}

Considering that our dataset was created using consistent parameters (same wavelength and antenna aperture size), the configuration cost is determined by the coordinates of the array elements. Each array is depicted as a set of coordinate pairs, featuring varying numbers of elements.

\subsection{Inputs}
The primary challenge in handling inputs arises from the neural network architecture selected. For architectures like the FNN used in our study, which are sensitive to the order and length of inputs, we address this issue by consistently ordering the antenna coordinates across all arrays. Additionally, to ensure uniform input length, we pad each input to match the maximum number of antennas in the dataset (1024).

Once the neural network provides a reliable estimate of the cost function, it forms the basis for optimizing antenna configurations. By structuring the configurations in a standardized input format, we enable the application of gradient-based optimization methods to refine antenna positions.

To optimize a specific antenna array configuration, we adjust the coordinates of the elements to find positions that minimize the cost function. Utilizing the neural network surrogate within a PyTorch framework \cite{paszke2019pytorch, paszke2017automatic}, we employ the Adam optimizer, a gradient-based optimization technique known for its computational efficiency and adjustable hyperparameters \cite{Kingma2014AdamAM}.

To enhance the optimization process further, we incorporate mechanisms that enforce physical constraints among the array elements. The first method is a minimum distance check, where each candidate configuration is evaluated at each iteration to ensure that elements maintain a set minimum distance constraint. If the candidate configuration satisfies this constraint, it is accepted, and the optimization advances to the next iteration. If it does not, the configuration reverts to the last valid arrangement, thereby concluding the optimization.

The second method involves adding a repulsion term as a penalty function within the cost model. This term imposes a penalty based on the proximity of element pairs, allowing the optimization to proceed uninterrupted. The penalty is designed to escalate significantly as any pair of elements nears the minimum distance threshold, actively discouraging configurations that breach spacing requirements. Designed to asymptotically approach infinity as the distance approaches the limit, this penalty can also be tailored to deter other undesirable behaviors, such as elements spreading too far apart or converging too closely, ensuring that all configurations remain within acceptable limits throughout the optimization process.

\subsection{Penalty Mechanism}
In our model, the penalty mechanism is flexible, allowing for the incorporation of custom penalties tailored to specific design requirements. These penalties are designed to tackle various constraints or optimization goals. The following example demonstrates the penalty function we utilized.

This function computes the Euclidean pairwise distance between each pair of elements, denoted as \(D_{ij}\) for elements \(i\) and \(j\), where \(i\) and \(j\) range from \(1, 2, \ldots, N_e\). A minimum threshold is also established, represented as \(\theta\).

\begin{equation}
P_{ij} = \log\left(\frac{1}{D_{ij} - \theta}\right) 
\end{equation}The total penalty \(P\) applied during the optimization is the aggregate of all individual penalties, scaled by a factor \(\epsilon\), which regulates the influence of the repulsion term on the optimization process:
\begin{equation}
    P = \epsilon \cdot \sum_{\substack{\forall i \in \text{Ant} \\ \forall j \in \text{Ant}}} P_{ij}
\end{equation}
Here, \(\text{Ant}\) denotes the set of elements in the configurations. The scalar \(\epsilon\) is crucial as a higher value increases the emphasis on maintaining distance between elements, effectively preventing them from approaching the minimum threshold.

The objective of the optimization process is to minimize the custom loss function defined as
\begin{equation}
    \text{loss} = f(\text{Ant}) + P
\end{equation}
where \(f(\text{Ant})\) represents the output of the surrogate cost network for a specific antenna configuration, and \(P\) is our custom penalty term.

\section{Results}

In this section, we showcase the performance improvements of the antenna arrays facilitated by the FNN and the Set Transformer, both with and without the implementation of the penalty mechanism designed to mitigate mutual coupling. Figure 3 displays the optimized sparse array generated by the FNN with the penalty mechanism incorporated. Figure 4 demonstrates the results for both the FNN and Set Transformer without the penalty mechanism, while Figure 5 shows the outcomes when the FNN is applied with and without the penalty mechanism. For simplicity, we chose the ten antenna array configurations with the lowest initial costs, with each optimization process taking approximately a minute to complete. Our method is readily scalable by executing multiple optimization processes in parallel.
\begin{figure}[htbp]
 \centerline{\includegraphics[width=\linewidth]{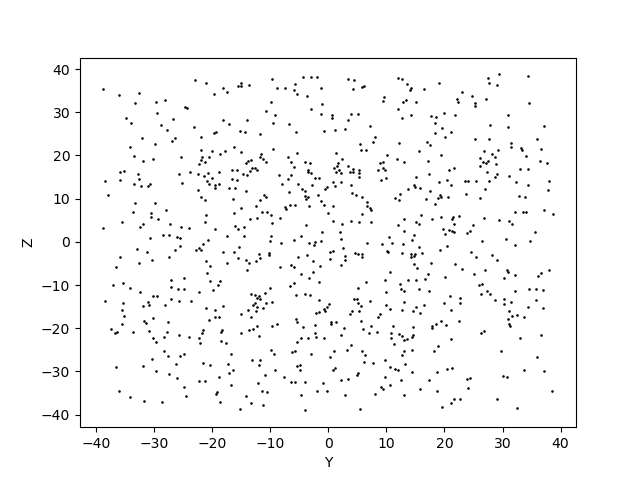}}
 \caption{Example of an optimized array configuration.}
 \label{fig:opt_coordinates}
\end{figure}

\begin{figure}[htbp]
 \centerline{\includegraphics[width=\linewidth]{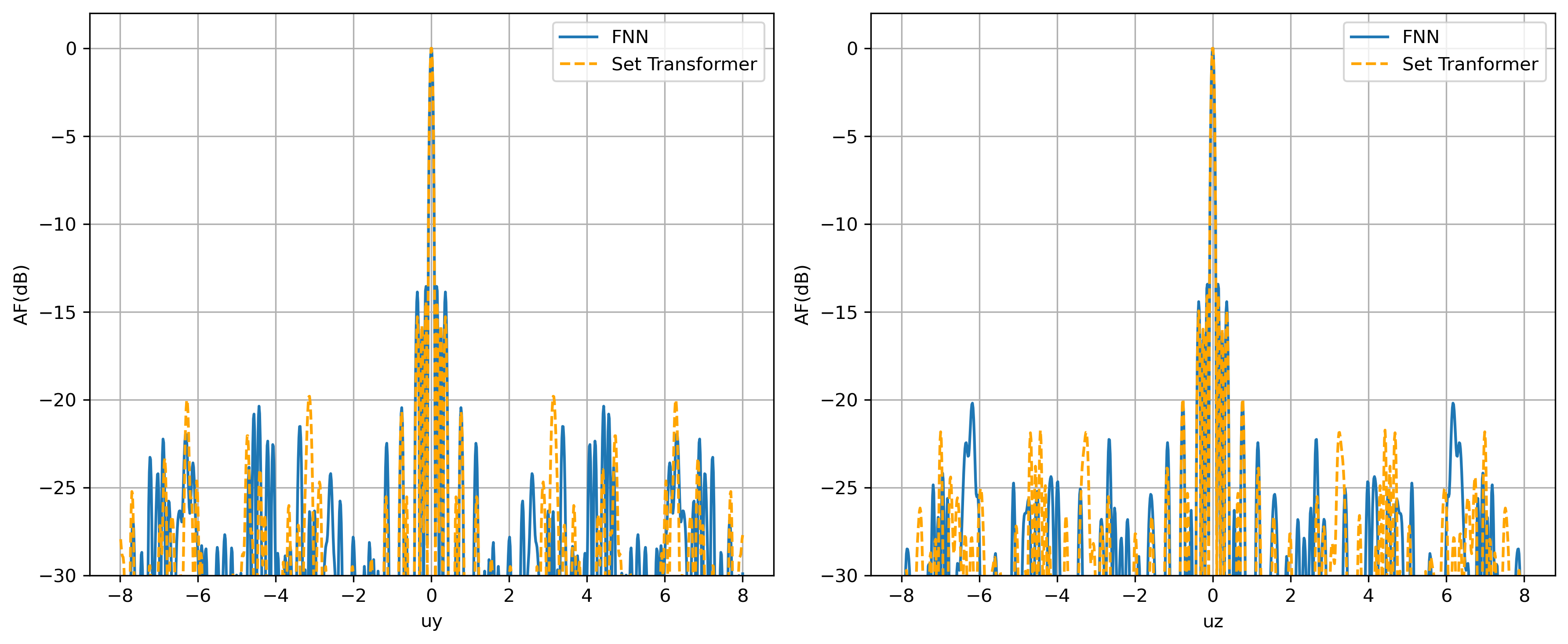}}
 \caption{The \( u_y \) and \(u_z\) cuts are derived using both the FNN and the Set Transformer. Solid lines indicate the results from the FNN, whereas dashed lines represent those obtained with the Set Transformer.}
 \label{fig:uy_cut}
\end{figure}

\begin{figure}[htbp]
\centerline{\includegraphics[width=\linewidth]{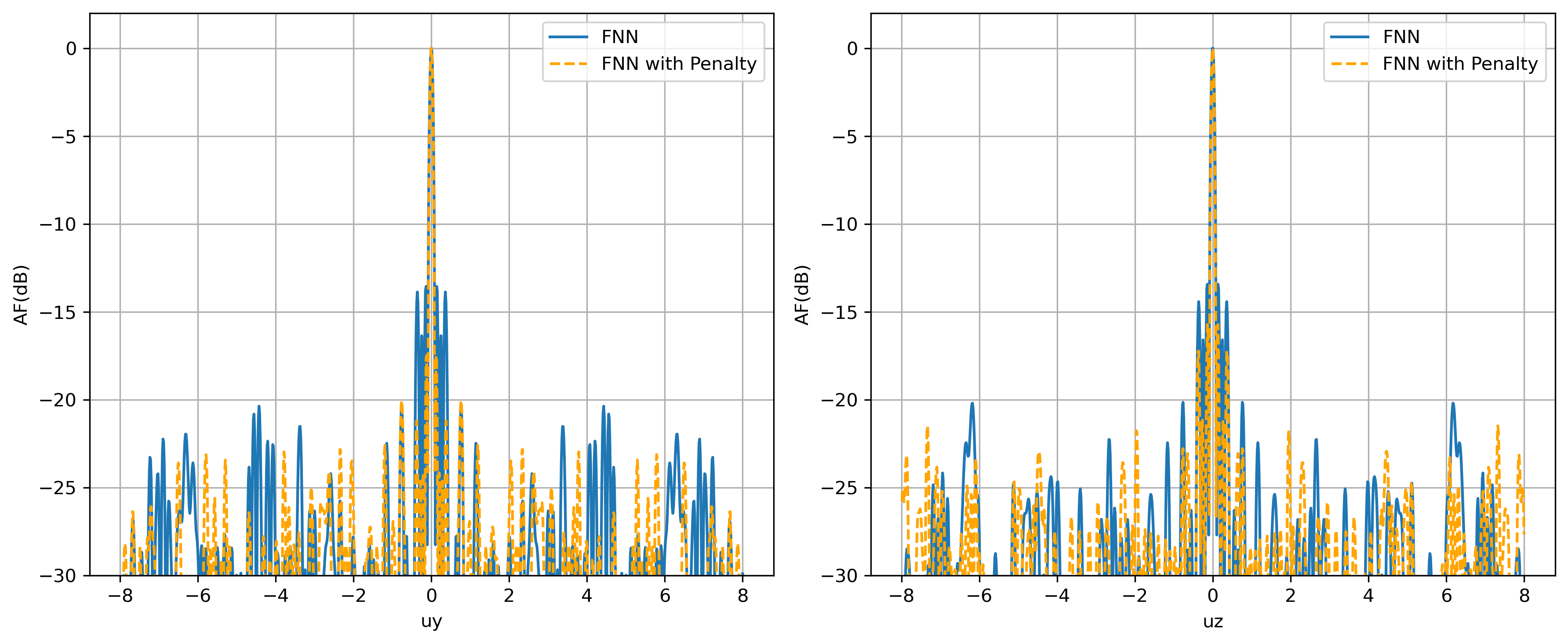}}
 \caption{The \( u_y \) and \(u_z\) cuts are derived using the FNN, both with and without the penalty mechanism. Solid lines indicate the results from the FNN without penalty, whereas dashed lines represent those obtained with penalty.}
 \label{fig:uy_cut2}
\end{figure}

The results indicate that the half-power beamwidth remains consistently similar (\(\zeta_\text{3dB} = 0.78^\circ\), where \(\zeta\) represents either \(EL\) or \(AZ\)) across all scenarios, including the FNN with and without the penalty mechanism, and the Set Transformer. In assessing the Side Lobe Level (SLL), we examine both the first and second SLL for a detailed comparison: the FNN with the penalty mechanism demonstrates the best performance, with the first SLL at (-17.46, -15.72) dB and the second SLL at (-20.03, -17.22) dB. Conversely, the Set Transformer and FNN without the penalty mechanism exhibit the first SLL at (-14.42, -14.06) dB and (-13.54, -13.42) dB respectively, and the second SLL at (-15.27, -14.93) dB and (-13.86, -14.41) dB, where the pair of values in parentheses represents the SLL measurements in \(u_y\) and \(u_z\) directions, respectively.

The FNN architecture incorporating the penalty mechanism demonstrates improved performance. A key approach to mitigating mutual coupling is to increase the minimum spacing between adjacent elements. With the penalty mechanism applied, the average minimum distance increases from $0.501\lambda$ (in the absence of the penalty) to $0.508\lambda$.

\begin{table}[!ht]
    \centering
    \caption{Comparison of Cost by Optimization without Penalty}
    \label{tab:w_pen_comp}
    \footnotesize 
    \begin{tabular}{|c|c|c|c|c|c|}
    \hline
    \multicolumn{2}{|c|}{} & \multicolumn{2}{c|}{FNN} & \multicolumn{2}{c|}{SetTrans.} \\ \hline
        Config. & Cost Bef. & Cost Aft. & \%Chg & Cost Aft. & \%Chg\\ \hline
        0 & -40463.98 & -65314.69 & -61.41\% & -75943.74 & -87.68\% \\ \hline
        1 & -39931.30 & -63976.30 & -60.21\% & -71023.64 & -77.86\% \\ \hline
        2 & -39767.27 & -64469.17 & -62.11\% & -68749.44 & -72.87\% \\ \hline
        3 & -39472.41 & -62825.44 & -59.16\% & -64694.44 & -63.89\% \\ \hline
        4 & -39431.64 & -65592.67 & -66.34\% & -65318.07 & -65.64\% \\ \hline
        5 & -39383.22 & -60226.57 & -52.92\% & -62768.85 & -59.37\% \\ \hline
        6 & -39036.94 & -60053.17 & -53.83\% & -68111.58 & -74.47\% \\ \hline
        7 & -38702.13 & -59404.24 & -53.49\% & -60284.31 & -55.76\% \\ \hline
        8 & -38690.57 & -64205.93 & -65.94\% & -58751.75 & -51.85\% \\ \hline
        9 & -38600.70 & -59656.36 & -54.54\% & -58166.98 & -50.68\% \\ \hline
    \end{tabular}
\end{table}
\subsection{Impact of Gradient Descent without Penalty}

An analysis of the 10 lowest cost configurations, conducted without the use of the custom penalty, revealed an average cost reduction of 59\%, with the most significant improvement observed at 66\%. A closer examination of the minimum distances between elements after optimization shows that these distances often closely approached the threshold (\(\theta = 0.5\)). This proximity to the threshold indicates that the optimization process aims to minimize costs, but it may compromise the physical feasibility and operational integrity of the antenna arrays by pushing the configurations to the limits of the minimum distance constraint.

\subsection{Effectiveness of the Penalty Function}

The application of gradient descent, incorporating a penalty for the minimum distance constraint, has shown substantial improvements in the cost of antenna array configurations. With our custom penalty mechanism, we observed an average improvement of 552\% across the configurations examined, with the largest improvement being 643\% for the FNN model. These results highlight the potential of gradient descent to effectively refine antenna array designs to achieve significant performance enhancements.

The use of a penalty function in optimization demonstrates its ability to guide the optimization process within the bounds of operational feasibility while still maximizing cost reduction. Optimization with a penalty ensures that improvements in antenna array configurations do not compromise practical applicability. Moreover, the variability introduced by adjusting \(\epsilon\) underscores the nuanced control that the penalty function offers over the optimization process. This control allows for a tailored balance between aggressive cost reduction and adherence to physical limitations. We found that \(\epsilon\) values of 12.5 for the FNN architecture and 1 for the Set Transformer architecture proved to be most optimal. These \(\epsilon\) values were determined by testing \(\epsilon\) in the range of 0.1 to 100.
\begin{table}[!ht]
    \centering
    \caption{Comparison of Cost by Optimization with Penalty}
    \label{tab:no_pen_comp}
    \footnotesize 
    \begin{tabular}{|c|c|c|c|c|c|}
    \hline
        \multicolumn{2}{|c|}{} & \multicolumn{2}{c|}{FNN} & \multicolumn{2}{c|}{SetTrans.} \\ \hline
        Config. & Cost Bef. & Cost Aft. & \%Chg & Cost Aft. & \%Chg \\ \hline
        0 & -40463.98 & -207003.25 & -411.57\% & -125854.84 & -211.02\% \\ \hline
        1 & -39931.30 & -253144.00 & -533.94\% & -74192.67 & -85.80\% \\ \hline
        2 & -39767.27 & -254191.15 & -539.19\% & -82944.35 & -108.57\% \\ \hline
        3 & -39472.41 & -240652.01 & -509.67\% & -82868.45 & -109.94\% \\ \hline
        4 & -39431.64 & -265697.50 & -573.81\% & -76439.64 & -93.85\% \\ \hline
        5 & -39383.22 & -265829.96 & -574.98\% & -81279.11 & -106.38\% \\ \hline
        6 & -39036.94 & -282136.09 & -622.74\% & -92020.96 & -135.72\% \\ \hline
        7 & -38702.13 & -234333.65 & -505.47\% & -83465.70 & -115.66\% \\ \hline
        8 & -38690.57 & -287802.15 & -643.85\% & -77825.16 & -101.14\% \\ \hline
        9 & -38600.70 & -273333.68 & -608.10\% & -66291.16 & -71.73\% \\ \hline
    \end{tabular}
\end{table}

\subsection{Comparative Analysis}

When evaluating the results of optimization methods, the significance of our surrogate cost function becomes evident, as well as the correlation between cost reduction and the minimal distance among components.

Optimization using a penalty function not only ensured a safer margin from critical thresholds but also surpassed the performance of optimization without such penalties. This enhancement was noticeable with both the FNN and Set Transformer models. This approach guarantees that configurations stay within safe operational boundaries, thus improving the practical effectiveness and efficiency of the optimized antenna arrays.

This comparative study highlights the necessity of adhering to key design constraints. Incorporating a specific penalty term within the optimization process was markedly more successful than methods lacking this feature. Using this tailored penalty term led to more substantial cost reduction and increased separation between array elements.

\section{Conclusion}
We have introduced a new design methodology for antenna arrays that utilizes deep learning to efficiently optimize large antenna configurations. This approach begins with designing sub-arrays as the fundamental building blocks for configuring the entire array. At the core of our method is the application of neural networks to drive subsequent refinements in array geometry. Using a deep learning approach, we have demonstrated an efficient way to optimize large antenna arrays, significantly reducing costs while adhering to physical constraints. Specifically, we reduced the costs of the ten antenna array configurations with the lowest initial cost by an average of 552\% with penalty and 59\% without penalty. Our methodology is general and can be applied to other arrays, including non-planar arrays. Researchers can use this novel approach in future work for applications such as phase tapering, beam shaping, and thinned array design, among others.

\bibliographystyle{ieeetr}
\bibliography{references}

\end{document}